\RequirePackage{fix-cm}
\documentclass[twocolumn]{svjour3}                     %
\smartqed  % flush right qed marks, e.g. at end of proof
\usepackage{graphicx}

\usepackage[utf8]{inputenc} % allow utf-8 input
\usepackage[T1]{fontenc}    % use 8-bit T1 fonts
\usepackage{hyperref}       % hyperlinks
\usepackage{url}            % simple URL typesetting
\usepackage{booktabs}       % professional-quality tables
\usepackage{amsfonts}       % blackboard math symbols
\usepackage{nicefrac}       % compact symbols for 1/2, etc.
\usepackage{microtype}      % microtypography
\usepackage{amsmath}
\usepackage{multirow}
\usepackage{graphicx}
\usepackage{xcolor}
\usepackage{epstopdf}
\usepackage{caption}
\usepackage{subcaption}
\usepackage{bm}
\usepackage{pgfplots}
\usepackage{footnote}
\usepackage{stfloats}
\usepackage{placeins}
\usepackage{color,soul}
\usepackage{dsfont}
\usepackage{amssymb}
\usepackage[boxruled]{algorithm2e}
\usepackage{pbox}
\usepackage{enumerate}
\usepackage{etextools}
\usepackage{mathtools}
\usepackage{tabulary}
\usepackage{dsfont}
\usepackage[titletoc]{appendix}

\newtheorem{thm}{Theorem} % reset theorem numbering for each chapter
\newtheorem{defn}[thm]{Definition} % definition numbers are dependent on theorem numbers

\newcommand{\changecolor}{black}

\title{Adversarial Noise Attacks of Deep Learning Architectures -- Stability Analysis via Sparse-Modeled Signals}

\author{Yaniv Romano$^*$ \thanks{Y. Romano and A. Aberdam contributed equally to this work.}
    \and
    Aviad Aberdam$^*$
    \and
	Jeremias Sulam
	\and
	Michael Elad \thanks{The research leading to these results has received funding from the Technion Hiroshi Fujiwara Cyber Security Research Center and the Israel Cyber Directorate. \newline Y. R. thanks the Zuckerman Institute, ISEF Foundation and the Viterbi Fellowship, Technion, for supporting this research.}
}

\institute{
            Y. Romano \at
              Department of Statistics, Stanford University\\
              \email{yromano@stanford.edu}
          \and
            A. Aberdam \at
              Electrical Engineering, Technion - Israel Institute of Technology \\
              \email{aaberdam@campus.technion.ac.il}
          \and
              J. Sulam \at
              Biomedical Engineering, Johns Hopkins University \\
              \email{jsulam@jhu.edu}           %  \\
           \and
           M. Elad \at
              Computer Science, Technion - Israel Institute of Technology \\
              \email{elad@cs.technion.com}
}

\usepackage{upgreek}

\def\x{{\mathbf x}}
\def\X{{\mathbf X}}
\def\v{{\mathbf v}}
\def\d{{\mathbf d}}

\def\Y{{\mathbf Y}}

\def\I{{\mathbf I}}

\def\D{{\mathbf D}}

\def\W{{\mathbf W}}
\def\w{{\mathbf w}}

\def\S{{\mathcal{S}}}

\def\PP{{\mathbf P}}
\def\RR{{\mathbb R}}
\def\SS{{\mathbf S}}

\def\E{{\mathbf E}}

\def\Gama{{\boldsymbol \Upgamma}}

\def\O{{\boldsymbol \Upomega}}
\def\lamda{{\boldsymbol \uplambda}}
\def\Eps{{\bm{\mathbf{\mathcal{E}}}}}

\def\Poe{{\text{P}_0^\Eps}}

\def\Poie{{\text{P}_{0,\infty}^\Eps}}

\def\DCPE{{\text{DCP}_\lamda^{\hspace{0.04cm} \Eps}}}

\def\pp{{\scriptscriptstyle{\PP}}}
\def\ss{{\scriptscriptstyle{\SS}}}

\newcommand{\norm}[1]{\left\lVert#1\right\rVert}
\newcommand{\abs}[1]{\left\lvert#1\right\rvert}

\usepackage{xcolor}

\mathtoolsset{showonlyrefs}

\date{Received: date / Accepted: date}
\pgfplotsset{compat=1.14
}
\begin{document}

\maketitle

\begin{abstract}
Despite their impressive performance, deep convolutional neural networks (CNN) have been shown to be sensitive to small adversarial perturbations. These nuisances, which one can barely notice, are powerful enough to fool sophisticated and well performing classifiers, leading to ridiculous misclassification results. In this paper we analyze the stability of state-of-the-art deep-learning classification machines to adversarial perturbations, where we assume that the signals belong to the (possibly multi-layer) sparse representation model. We start with convolutional sparsity and then proceed to its multi-layered version, which is tightly connected to CNN. Our analysis links between the stability of the classification to noise and the underlying structure of the signal, quantified by the sparsity of its representation under a fixed dictionary.
In addition, we offer similar stability theorems for two practical pursuit algorithms, which are posed as two different deep-learning architectures  -- the Layered Thresholding and the Layered Basis Pursuit. Our analysis establishes the better robustness of the later to adversarial attacks. We corroborate these theoretical results by numerical experiments on three datasets: MNIST,
CIFAR-10 and CIFAR-100. 
% we numerically compare robustness to adversarial noise between the layered thresholding (CNN) and the proposed layered basis pursuit architectures on three datasets: MNIST, CIFAR-10 and CIFAR-100, demonstrating the advantage of the later as reflected in our analysis. 
% Furthermore, our claims can be translated to a practical regularization term that provides a new interpretation to the robustness of Parseval Networks. Also, the proposed theory justifies the increased stability of the recently emerging layered basis pursuit architectures, when compared to the classic forward-pass.
\end{abstract}

\section{Introduction}

Deep learning, and in particular Convolutional Neural Networks (CNN), is one of the hottest topics in data sciences as it has led to many state-of-the-art results spanning across many domains \cite{lecun2015deep,goodfellow2016deep}. Despite the evident great success of classifying images, it has been recently observed that CNN are highly sensitive to adversarial perturbations in the input signal \cite{szegedy2013intriguing,goodfellow2014explaining,liu2016delving}. An \emph{adversarial example} is a corrupted version of a valid input (i.e., one that is classified correctly), where the corruption is done by adding a perturbation of a small magnitude to it. This barely noticed nuisance is designed to fool the classifier by maximizing the likelihood of an incorrect class. This phenomenon reveals that state-of-the-art classification algorithms are highly sensitive to noise, so much so that even a single step in the direction of the sign of the gradient of the loss function creates a successful adversarial example \cite{goodfellow2014explaining}. Furthermore, it has been shown that adversarial examples that are generated to attack one network are powerful enough to fool other networks of different architecture and database \cite{liu2016delving}, being the key to the so-called ''black-box" attacks that have been demonstrated in some real-world scenarios \cite{kurakin2016adversarial}.

Adversarial training is a popular approach to improve the robustness of a given classifier \cite{goodfellow2014explaining}. It aims to train a robust model by augmenting the data with adversarial examples generated for the specific model and/or transferred from other models. Preprocessing \cite{liao2017defense} is another defense strategy, suggesting to denoise the input signal first, and then feed this purified version of the signal to the classifier. Indeed, the above defense methods improve the stability of the network; however, these are trained based on adversarial examples that are generated in specific ways. It is quite likely that future work could offer a different generation of adversarial examples that question again the reliability and robustness of such given networks. %{We note that there are other ways to improve the stability of the network, such as robust optimization \cite{shaham2015understanding}, using distillation \cite{papernot2016distillation}, and more.}

In this paper we provide a principled way to analyze the robustness of a classifier using the vast theory developed in the field of sparse representations. We do so by analyzing the classifier's robustness to adversarial perturbations, providing an upper bound on the permitted energy of the perturbation, while still safely classifying our data. The derived bounds are affected by the classifier's properties and the structure of the signal. Our analysis assumes that the signals of interest belong to the sparse representation model, which is known for its successful regression and classification performance \cite{Mairal2014,Elad_Book}, and was recently shown to be tightly connected to CNN \cite{Papyan2017convolutional}. We commence by analyzing a shallow convolutional sparse model and then proceed to its multi-layer extension.
More concretely, suppose we are given a clean signal that is assigned to the correct class. How much noise of bounded energy can be added to this signal and still guarantee that it would be classified accurately? Our work shows that the bound on the energy of the noise is a function of the sparsity of the signal and the characteristics of the dictionaries (weights).

We proceed by considering specific and practical pursuit algorithms that aim to estimate the signal's representations in order to apply the classification. Our work investigates two such algorithms, the non-negative Layered Thresholding (L-THR), which amounts to a conventional feed-forward CNN, and the non-negative Layered Basis-Pursuit (L-BP), which is reminiscent of an RNN (Residual Neural Network) architecture. Our analysis exposes the ingredients of the data model governing the sensitivity to adversarial attacks, and clearly shows that the later pursuit (L-BP) is more robust. 

The bounds obtained carry in them practical implications. More specifically, our study indicates that a regularization that would take the dictionaries' coherence into account can potentially improve the stability to noise. 
Interestingly, a regularization that aligns well with our findings was tested empirically by Parseval Networks \cite{moustapha2017parseval} and indeed shown to improve the classification stability. As such, one can consider our work as a theoretical explanation for the empirical success of \cite{moustapha2017parseval}. Another approach that is tightly connected to our analysis is the one reported in \cite{Zeiler2010,mahdizadehaghdam2018deep}. Rather than relying on a simple L-THR, these papers suggested solving a variant of the L-BP algorithm, in an attempt to promote sparse feature maps. Interestingly, it was shown in \cite{mahdizadehaghdam2018deep} that the ''fooling rate" in the presence of adversarial perturbation is significantly improved, serving as another empirical evidence to our theoretical conclusions. As will be shown in this paper, promoting sparse solutions and incoherent dictionaries is crucial for robust networks, as evidenced empirically in the above two papers \cite{moustapha2017parseval,mahdizadehaghdam2018deep}. 

\textcolor{\changecolor}{We should note that this work does not deal with the learning phase of the networks, as we assume that we have access to the true model parameters. Put on more practical terms, our work analyzes the sensitivity of the chosen inference architectures to 
malicious noise, by imposing assumptions on the filters/dictionaries and the incoming signals. These architectures follow the pursuit algorithms we explore, and their parameters are assumed to be known, obtained after learning.}

\textcolor{\changecolor}{Moving to the experimental part, we start by demonstrating the derived theorems on a toy example, in order to better clarify the message of this work. 
Our simulations carefully illustrate how the L-BP is more stable to adversarial noise, when compared with the regular feed-forward neural network (i.e., the L-THR), and this is shown both in theoretical terms (showing the actual bounds) and in empirical performance.  In order to further support the theoretical claims made in this paper, we numerically explore the stability of the L-THR and the L-BP architectures on actual data and learned networks. Note that in these experiments the theoretical assumptions do not hold, as we do not have an access to the true model. In this part} we consider three commonly tested datasets: MNIST \cite{lecun2010mnist}, CIFAR-10 \cite{krizhevsky2014cifar} and CIFAR-100 \cite{krizhevsky2014cifar}. Our experiments show that the L-BP is indeed more robust to noise attacks, where those are computed using the Fast Gradient Sign Method (FGSM) \cite{goodfellow2014explaining}. 

This paper is organized as follows: In Section \ref{sec:CSCD} we start by reviewing the basics of the convolutional sparse coding model and then proceed to its multi-layered version, which is tightly connected to CNN. Then, using Sparseland tools we establish a connection between the stability of the classification to adversarial noise and the underlying structure of the signal, quantified by the sparsity of its representation. We commence by analyzing shallow networks in Section \ref{sec:Shallow} and then continue to deeper settings in Section \ref{sec:MLCSC}. In addition, we offer similar stability theorems for two pursuit algorithms, which are posed as two different deep-learning architectures -- the L-THR and the L-BP. In Section \ref{sec:numerical} we numerically study the stability of these architectures demonstrating the theoretical results, {\color{\changecolor} starting with a toy example using simulated data, and then continuing with tests on real data.} We conclude in Section \ref{sec:conclusions} by delineating further research directions.

\section{Background and problem setup}
\label{sec:CSCD}

Consider a set $ \left\{s^j\right\}_{j} = \left\{\left(\X^j, y^j\right)\right\}_{j} $ of high dimensional signals $ \X^j \in \mathcal{X}\subseteq \RR^N $ and their associated labels $ y^j \in \mathcal{Y} $. Suppose that each signal $ \X^j = \D \Gama^j $ belongs to the (possibly multi-layer convolutional \cite{Papyan2017convolutional}) sparse representation model, where $ \D $ is a dictionary and $ \Gama^j $ is a sparse vector. Suppose further that we are given a linear classifier that operates on the \emph{sparse representation} $ \Gama^j $ and successfully discriminates between the different classes. 

\begin{figure}[h]
	\centering
	\includegraphics[width=0.65\linewidth]{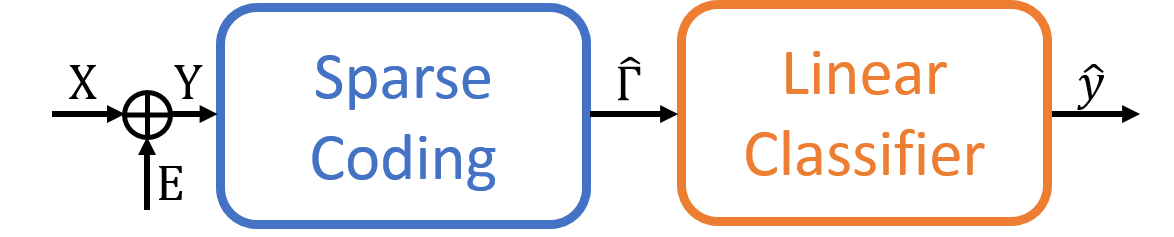}
	\caption{\textcolor{\changecolor}{The classification scheme consists of a sparse coding block and a linear classifier. The adversarial noise $\E$ aims to fail the classification, $\hat{y}\ne y$, while having of the smallest possible energy.}}
	\label{Fig:SingleLayerModel}
\end{figure}

Ignoring the superscript $ j $ for clarity, given the input $ s = (\X, y) $ the adversary's goal is to find an example $ \Y = \X + \E $, such that the energy of $ \E $ is small, and yet the model would misclassify $ \Y $. \textcolor{\changecolor}{Figure \ref{Fig:SingleLayerModel} depicts this classification scheme.} We consider the class of $ \ell_p $ bounded adversaries, in the sense that for a given energy $ \epsilon $, the adversarial example satisfies \mbox{$ \|\Y - \X \|_p = \|\E\|_p \leq \epsilon $}.

How much perturbation $ \E \in \RR^N $ of bounded energy $ \epsilon $ can be added to $ \X $ so as the measurement $ \Y = \X + \E $ will still be assigned to the correct class? What is the effect of the sparsity of the true representation? What is the influence of the dictionary $ \D $ on these conclusions? How can we design a system that will be robust to noise based on the answers to the above questions? These questions are the scope of this paper. Before addressing these, in this section we provide the necessary background on several related topics.

\subsection{Convolutional sparse coding}

\begin{figure*}[t]
	\centering
	\includegraphics[width=1\textwidth]{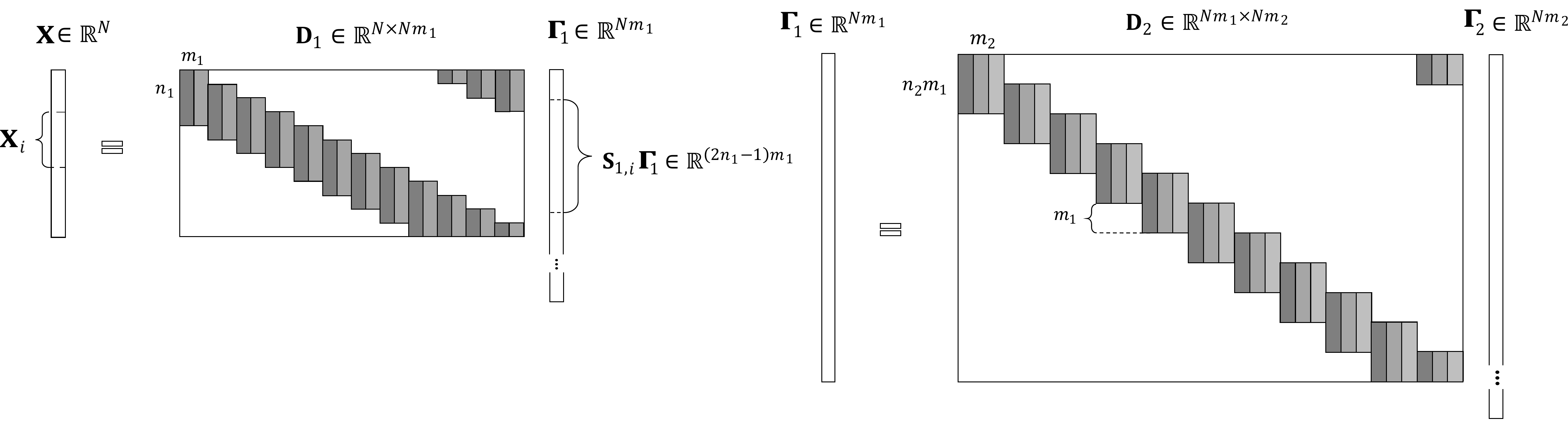}
	\caption{Left: The global convolutional system $\X=\D_1\Gama_1$, along with the representation of the $i$-th patch $\SS_{1,i} \Gama_1$. Right: The second layer of the multi-layer CSC model, given by $ \Gama_1 = \D_2 \Gama_2 $.}
	\label{Fig:LocalSystem}
\end{figure*}

The convolutional sparse coding (CSC) model assumes that a signal $ \X \in \RR^{N} $ can be represented as $ \X = \D \Gama $, where $ \D\in\RR^{N\times Nm} $ is a given convolutional dictionary and $ \Gama \in \RR^{Nm} $ is a sparse vector. The dictionary $ \D $ is composed of $ m $ local unique filters of length $ n $, where each of these is shifted at every possible location (see Figure \ref{Fig:LocalSystem} left, here ignore the subscript '1' for clarity). The special structure of this matrix implies that the $ i $-th patch $ \x_i \in \RR^n $ extracted from the global signal $ \X $ has an underlying shift-invariant local model \cite{Papyan2017working}. Concretely, $ \x_i = \O\SS_i\Gama $, where $ \O $ is a fixed matrix shared by all the overlapping patches, multiplied by the corresponding stripe vector $ \SS_i\Gama_i \in \RR^{(2n - 1)m} $, where $ \SS_i \in \RR^{(2n - 1)m \times  mN }$ extracts the stripe from the global $ \Gama $. 

Building upon the local structure of this model, it was shown in \cite{Papyan2017working} that measuring the local sparsity of $ \Gama $ rather than the global one is much more informative. The notion of local sparsity is defined by the $ \ell_{0,\infty} $ pseudo-norm, expressed by
$ \|\Gama\|_{0,\infty}^\ss = \max_i \|\SS_i\Gama\|_0, $
which counts the maximal number of non-zeros in the stripes (and hence the superscript $ \ss $) of length $(2n - 1)m$ extracted from $ \Gama $.

In the setting of this paper, we are given a noisy measurement of $ \X = \D\Gama $, formulated as $ \Y = \X + \E $, where $\E$ is an $ \ell_p $-bounded \emph{adversarial perturbation}. In the $ \ell_2 $ case, the pursuit problem of estimating $ \Gama $ given $ \Y,\D $ and the energy of $ \E $ (denoted by $ \epsilon $) is defined as
\begin{align} \label{Eq:CSCpursuit}
\quad (\Poie): \quad \underset{\Gama}{\min} \ \|\Gama\|_{0,\infty}^\ss \ \text{ s.t. }\ \|\Y-\D\Gama\|^2_2\leq\epsilon^2.
\end{align}
The stability of the above problem and practical algorithms (Orthogonal Matching Pursuit -- OMP, and Basis Pursuit -- BP) that aim to tackle it were analyzed in \cite{Papyan2017working}. Under the assumption that $ \Gama $ is ``locally sparse enough'', it was shown that one can obtain an estimate $ \hat{\Gama} $ that is close to the true sparse vector $ \Gama $ in an $ \ell_2 $-sense. The number of non-zeros in $ \Gama $ that guarantees such a stable recovery is a function of $ \epsilon $ and the characteristics of the convolutional dictionary $ \D $.

Two measures that will serve us in our later analysis are (i) the extension of the Restricted Isometry Property (RIP) \cite{candes2008restricted} to the convolutional case, termed SRIP \cite{Papyan2017working}, and (ii) the mutual coherence. The SRIP of a dictionary $ \D $ of cardinality $ k $ is denoted by $ \delta_k $. It measures how much the multiplication of a locally sparse vector $ \v, \|\v\|_{0,\infty}^\ss = k $ by $\D$ changes its energy (see definition 14 in \cite{Papyan2017working}). A small value of $ \delta_k (\ll 1) $ implies that $ \D $ behaves almost like an orthogonal matrix, i.e. $ \|\D\v\|_2 \approx \|\v\|_2 $.

The second measure that we will rely on is the mutual coherence of a dictionary with $ \ell_2 $ normalized columns, which is formulated as $ \mu(\D) = {\max}_{i\neq j} \ |\d_i^T \d_j|, $ where $ \d_j $ stands for the $ j $-th column (atom) from $ \D $. In words, $ \mu(\D) $ is the maximal inner product of two distinct atoms extracted from $ \D $.

%An example to such algorithms are the Orthogonal Matching Pursuit (OMP), Basis Pursuit (BP), and the Thresholding algorithm.

\subsection{Multi-layer CSC}

The multi-layer convolutional sparse coding (ML-CSC) model is a natural extension of the CSC to a hierarchical decomposition. Suppose we are given a CSC signal $ \X = \D_1\Gama_1 $, where $ \D_1 \in \RR^{N \times N m_1} $ is a convolutional dictionary and $ \Gama_1 \in \RR^{Nm_1}$ is the (local) sparse representation of $ \X $ over $ \D_1 $ (see Figure \ref{Fig:LocalSystem} left). The ML-CSC pushes this structure forward by assuming that the representation itself is structured, and can be decomposed as $ \Gama_1 = \D_2 \Gama_2 $, where $ \D_2 \in \RR^{Nm_1 \times Nm_2} $ is another a convolutional dictionary, multiplied by the locally sparse vector $ \Gama_2 \in \RR^{Nm_2}$ (see Figure \ref{Fig:LocalSystem} right). Notice that $ \Gama_1 $ has two roles, as it is the representation of $ \X $, and a signal by itself that has a CSC structure. The second dictionary $ \D_2 $ is composed of $ m_2 $ local filters that skip $ m_1 $ entries at a time, where each of the filters is of length $ n_2m_1 $. This results in a convolution operation in the spatial domain of $ \Gama_1 $ but not across channels ($ \Gama_1 $ has $ m_1 $ channels), as in CNN. The above construction is summarized in the following definition (Definition 1 in  \cite{Papyan2017convolutional}):
\begin{defn} \label{Def:DCP}
	For a global signal $\X$, a set of convolutional dictionaries $\{ \D_i \}_{i=1}^K$, and a vector $\lamda$, define the ML-CSC model as:
	\begin{align}
	\Gama_{i-1} = \D_i \Gama_i, \ \ \| \Gama_i \|_{0,\infty}^\ss \leq \lambda_i \quad \forall \ 1 \leq i \leq K
	\end{align}
	where $ \Gama_{0} = \X $, and the scalar $\lambda_i$ is the $i$-th entry in $\lamda$.
\end{defn}

Turning to the pursuit problem in the noisy regime, an extension of the CSC pursuit (see Equation \eqref{Eq:CSCpursuit}) to the multi-layer setting (of depth $ K $) can be expressed as follows:
\begin{defn} (Definition 2 in \cite{Papyan2017convolutional})
	For a global signal $\Y$, a set of convolutional dictionaries $\{ \D_i \}_{i=1}^K$, sparsity levels $\lamda$ and noise energy $\epsilon$, the deep coding problem is given by
	\begin{align*}
	\hspace{0.2cm} (\DCPE): &  \text{ find }  \{\Gama_i\}_{i=1}^{K} \\ 
	& \text{ s.t. } 
	\| \Y   - \D_1 \Gama_1\|_2  \leq \epsilon,       & \\
	&\qquad \Gama_{i-1}      = \D_i \Gama_i,&  \\
	&\qquad \| \Gama_i \|_{0,\infty}^\ss	 \leq \lambda_i,  & \forall \ 1 \leq i \leq K.
	\end{align*}
\end{defn}

How can one solve this pursuit task? The work reported in \cite{Papyan2017convolutional} has shown that the forward pass of CNN is in fact a pursuit algorithm that is able to estimate the underlying representations $ \Gama_1, \dots, \Gama_K $ of a signal $ \X $ that belongs to the ML-CSC model. Put differently, the forward pass was shown to be nothing but a non-negative layered thresholding pursuit, estimating the representations $ \Gama_i $ of the different layers. To better see this, let us set  $ \hat{\Gama}_0 = \Y $ and define the classic thresholding pursuit \cite{Elad_Book}, $ \hat{\Gama}_i = \S^+_{\beta_i}(\D_i^T\hat{\Gama}_{i-1}) $, for $ 1 \leq i \leq K $. The term $ \D_i^T\hat{\Gama}_{i-1} $ stands for convolving $ \hat{\Gama}_{i-1} $ (the feature map) with the filters of $ \D_i $ (the weights), and the soft non-negative thresholding function $ \S^+_{\beta_i}(\v) = \max\{0,\v - \beta_i\} $ is the same as subtracting a bias $ \beta_i $ from $ \v $ and applying a ReLU nonlinearity. In a similar fashion, the work in \cite{Papyan2017convolutional} offered to replace the Thresholding algorithm in the sparse coding blocks with Basis-Pursuit, exposing a recurrent neural network architecture that emerges from this approach.

\begin{figure}[h]
	\centering
	\includegraphics[width=0.95\linewidth]{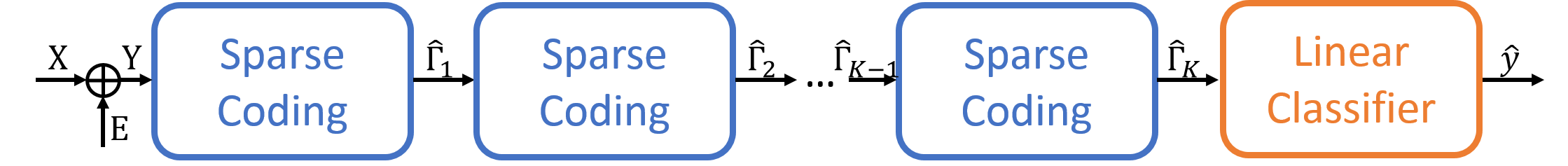}
	\caption{\textcolor{\changecolor}{A deep classification scheme consisting of a chain of sparse coding blocks and a linear classifier.}}
	\label{Fig:MultiLayerModel}
\end{figure}

This connection of CNN to the pursuit of ML-CSC signals was leveraged \cite{Papyan2017convolutional} to analyze the stability of CNN architectures. Their analysis concentrated only on the feature extraction stage -- the pursuit -- and ignored the classification step and the role of the labels. In this paper we build upon this connection of Sparseland to CNN, and extend the analysis to cover the stability of layered pursuit algorithms when tackling the classification task in the presence of noise. \textcolor{\changecolor}{In Section \ref{sec:MLCSC} we shall consider a classifier consisting of a chain of sparse coding blocks and a linear classifier at its deepest layer, as depicted in Figure \ref{Fig:MultiLayerModel}. Our work aims to analyze the stability of such a scheme, suggesting that replacing the pursuit algorithm in the sparse coding blocks from Thresholding to Basis-Pursuit yields a more stable architecture with respect to adversarial noise, both theoretically and practically.}

\begin{figure*}[t!]
	\centering
	\begin{subfigure}{1\linewidth}
	    \centering \includegraphics[width=0.75\linewidth]{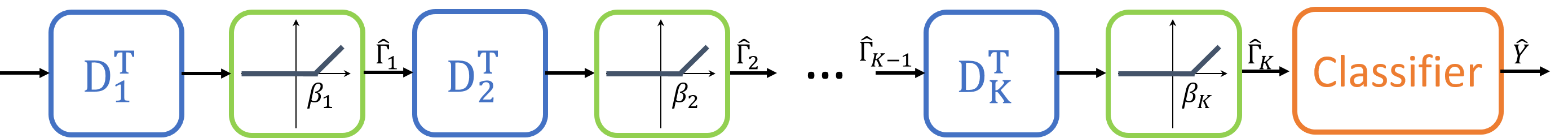}
    	\caption{\textcolor{\changecolor}{The Layered-Thresholding classifier (L-THR), corresponding to a CNN - a feed-forward convolutional neural network.}}
    	\label{Fig:LTHR}
	\end{subfigure}
	\begin{subfigure}{1\linewidth}
	    \centering \includegraphics[width=0.75\linewidth]{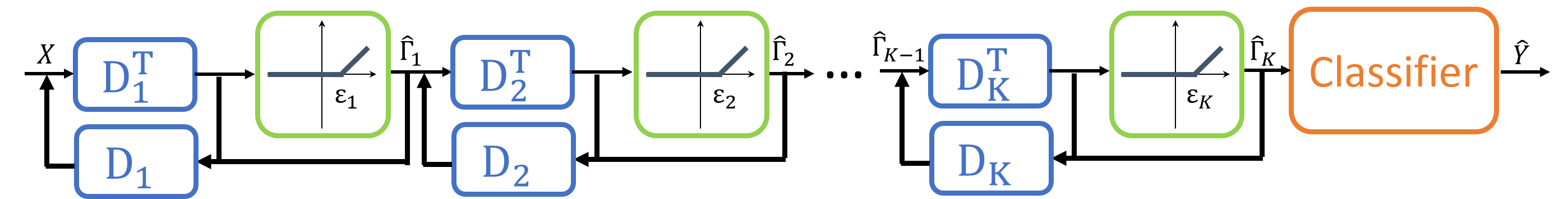}
    	\caption{\textcolor{\changecolor}{The Layered-Basis-Pursuit classifier (L-BP), corresponding to a CNN with additional feedback loops.}}
    	\label{Fig:LBP}
	\end{subfigure}
	\caption{\textcolor{\changecolor}{The deep classifier architectures considered in this work.}}
\end{figure*}

{\color{\changecolor}
More specifically, we first study the stability to adversarial noise of the feed-forward CNN classifier. Or equivalently, where each of the pursuit algorithms in 
Figure \ref{Fig:MultiLayerModel} is chosen to be the Thresholding. This architecture is depicted in Figure \ref{Fig:LTHR}. Then we switch to the Basis-Pursuit as the sparse coding, serving better the sparse model, and resulting a new deep-learning architecture with the same number of parameters but with additional feedback loops as illustrated in Figure\footnote{\textcolor{\changecolor}{Note that in this scheme, the number of iterations for each BP pursuit stage is implicit, hidden by the number of loops to apply. More on this is given in later sections.}} \ref{Fig:LBP}. We now give more formal definitions of these two schemes.}

\textcolor{\changecolor}{
\begin{defn}\label{Thm:LTHR}(L-THR)
	For an ML-CSC signal $\Y = \X + \E$ with convolutional dictionaries $ \{\D_i\}_{i=1}^K $, thresholds $\{\beta_i\}_{i=1}^K$, and a classifier $ (\w, \omega) $, define the Layered-Thresholding (L-THR) algorithm as: Apply  
	\begin{eqnarray*} 
	     ~~~\hat{\Gama}_i = \S^+_{\beta_i}(\D_i^T\hat{\Gama}_{i-1})~~\text{for}~~ i=1,2,~\ldots~, K, 
	\end{eqnarray*}
	and assign $y = sign\left(f(\hat{\Gama}_K)\right)$, where 
	\begin{eqnarray*} 
    	~~~f(\hat{\Gama}_K) = \w^T \hat \Gama_K + \omega.
	\end{eqnarray*}
\end{defn}}

\textcolor{\changecolor}{
\begin{defn}\label{Thm:LBP}(L-BP)
	For an ML-CSC signal $\Y = \X + \E$ with convolutional dictionaries $ \{\D_i\}_{i=1}^K $, Lagrangian multipliers $\{\epsilon_i\}_{i=1}^K$, and a classifier $ (\w, \omega) $, define the Layered-Basis-Pursuit (L-BP) algorithm as: Apply 
	\begin{eqnarray*}
	    ~~~\hat{\Gama}_{i} = \underset{\Gama_i}{\arg\min} \ \xi_i \|\Gama_i\|_1 + \frac{1}{2}\| \D_i\Gama_i - \hat{\Gama}_{i-1} \|_2^2\\ 
	    ~~~~~~~~\text{for}~~ i=1,2,~\ldots~, K,
	 \end{eqnarray*}
	 and assign $y = sign\left(f(\hat{\Gama}_K)\right)$, where 
	\begin{eqnarray*} 
    	~~~f(\hat{\Gama}_K) = \w^T \hat \Gama_K + \omega.
	\end{eqnarray*}
\end{defn}}

\section{First steps: shallow sparsity}
\label{sec:Shallow}
\subsection{Two-class (binary) setting} \label{sec:Two-Class}
Herein we consider the a binary classification setting (i.e. $ \mathcal{Y} = \{1,-1\} $) in which \emph{a linear classifier is given to us, being part of the generative model}. This classifier is defined by the couple $ \left(\w,\omega\right) $, where $ \w \in \RR^{Nm} $ is a weight vector and $ \omega $ is a bias term (a scalar).
%We assume that when the classifier operates on the true representation $ \Gama $ of $ \X $ it \emph{perfectly} discriminates between the two possible classes.
Put formally, the model we shall study in this subsection is given by the definition below.
\begin{defn} \label{def:binaryClean}
	A convolutional Sparseland signal $ \X = \D\Gama, \ \|\Gama\|_{0,\infty}^\ss \leq k $ is said to belong to class $ y = 1 $ when the linear discriminant function $ f\left(\Gama\right) = \w^T\Gama + \omega $ satisfies $ f(\Gama) > 0 $, and $ y = -1 $ otherwise.
\end{defn}
\noindent The expression $ \w^T\Gama + \omega $ defines a linear discriminant function for which the decision surface $ f(\Gama') = 0 $ is a hyperplane in the feature domain (but not in the signal domain), where $ \Gama' $ is a point that lies on the decision boundary. As such, one can express the distance from the decision boundary \emph{in the feature domain} as
$ \mathcal{O}_{\mathcal{B}}(\X,y) = y f(\Gama) = y\left(\w^T\Gama + \omega\right),
 $ where the subscript $ {\mathcal{B}} $ stands for Binary.
Notice that the larger the value of $ \mathcal{O}_{\mathcal{B}}(\X,y) $, the larger the distance to the decision boundary, as it is defined by $ \mathcal{O}_{\mathcal{B}}(\X',y) = 0$. Following this rational, $ \mathcal{O}_{\mathcal{B}}(\X,y) $ is often termed the \emph{score} or the \emph{output margin}. The measure $ \mathcal{O}_{\mathcal{B}}(\X,y) $ is $ \X $-dependent, and thus we have an interest in its extreme value,
\begin{align}
\mathcal{O}_{\mathcal{B}}^* = \underset{\{\X^j,y^j\}_{j}}{\text{min}} \ \mathcal{O}_{\mathcal{B}}(\X^j,y^j),
\end{align}
being a property of our data and the classifier's parameters, making our claims universal and not signal-specific. As will be shown hereafter, classification robustness is directly related to this quantity. Moreover, we emphasize that there are two margins involved in our analysis: (i) the above-described input data margin, which cannot be controlled by any learning scheme, and (ii) the margin that a classifier obtains when operating on a perturbed input signal, resulting in the evaluated representation $ {\hat \Gama} $. The results in this work rely on these two measures, as we aim to make sure that the former margin is not diminished by the practical classifier design or the adversarial noise.
%In addition, as we will see next, as otherwise the classifier at hand is highly sensitive to noise, and in particular to adversarial one.

This takes us naturally to the adversarial setting. The problem we consider is defined as follows:
\begin{defn}
	For a signal $ \Y = \X + \E $ with a true label $ y $, a convolutional dictionary $ \D $, a perturbation energy $ \epsilon $, and a classifier $ (\w,\omega) $, define the binary classification algorithm as:
	\begin{align*}
	\text{Solve} \qquad & \hat{\Gama} = \arg \min_{\Gama} \ \| \Gama \|_{0,\infty}^\ss \\
	\text{ s.t. } \qquad &
	\| \Y - \D \Gama\|_2 \leq \epsilon \\
	\text{and assign } &  \hat{y} = sign\left(f(\hat{\Gama})\right)
	\end{align*}
	where $ f(\hat{\Gama}) = \w^T \hat \Gama + \omega $.
\end{defn}
\noindent Notice that the signal is assigned to the correct class if $ sign(f(\Y)) = y$, or, equivalently when $ \mathcal{O}_{\mathcal{B}}(\Y,y) = y f(\Y) = y (\w^T \hat{\Gama} + \omega) > 0 $.
In words, the pursuit/sparse-coding step projects the perturbed signal $ \Y $ onto the model by estimating the representation $ \hat{\Gama} $, which in turn is fed to a classifier as formulated by $ f(\Y) $. 
In the remaining of this paper we shall study the conditions on $ \X,\D $ and $ \epsilon $ which guarantee that $ \mathcal{O}_{\mathcal{B}}(\Y,y) > 0$, i.e., the input signal $ \Y $ will be assigned to the correct class despite of the adversarial perturbation and the limitations of a specific pursuit algorithm.
The model assumptions that we are considering allow us to reveal the underlying characteristics of the signal (e.g. the properties of the dictionary and the sparsity level) that are crucial for a successful prediction. Put differently, \emph{we aim to reveal the ingredients that one should consider when designing a robust classification system. Notice that we concentrate on the inference stage only and do not analyze the learning part.} In fact, similarly to \cite{fawzi2018analysis}, we see this as an advantage since it keeps the generality of our findings.

Let us start our discussion by first studying the stability of the binary classification algorithm to noise:
\begin{thm}{(Stable Binary Classification of the CSC Model):} \label{Thm:StablityRIP}
	Suppose we are given a CSC signal $ \X = \D\Gama $, $ \|\Gama\|_{0,\infty}^\ss \leq k $, contaminated with perturbation $ \E $ to create the signal $ \Y = \X + \E $, such that $ \|\E\|_{2} \leq \epsilon $. Suppose further that $ \mathcal{O}_{\mathcal{B}}^* > 0 $ and denote by $\hat{\Gama}$ the solution of the $ \Poie $ problem (see Equation \eqref{Eq:CSCpursuit}). Assuming that
	$ \delta_{2k} < 1 - \left(\frac{2{\left\|\w\right\|_2}\epsilon}{\mathcal{O}_{\mathcal{B}}^*}\right)^2, $
	then $ sign(f(\Gama)) = sign(f(\hat{\Gama}))$.
	
	Considering the more conservative bound that relies on $ \mu(\D) $, and assuming that 
    \begin{equation*}
        \| \Gama \|_{0,\infty}^{\ss} < k = \frac{1}{2} \left(1 + \frac{1}{\mu(\D)}\left[1 - \left(\frac{2{\left\|\w\right\|_2}\epsilon}{\mathcal{O}_{\mathcal{B}}^*}\right)^2\right]\right),
    \end{equation*}
    then $ sign(f(\Gama)) = sign(f(\hat{\Gama}))$.  
\end{thm}
The proof of this theorem and the remaining ones are given in the Appendix. Among various implications, the above theorem shows the effect of $ \D $ and its properties on the stability of the classifier. A dictionary with $ \delta_{2k} \ll 1 $ tends to preserve the distance between any pair of \emph{locally} $ k $-sparse vectors (defined by the $ \ell_{0,\infty} $ norm), which turns to be crucial for robust classification. The benefit of switching from the SRIP to $ \mu(\D) $ is that the latter is trivial to compute, but with the cost of weakening the result. The expected and somewhat unsurprising conclusion of Theorem \ref{Thm:StablityRIP} is that the score of the classifier plays a key role for a stable classification -- the larger the distance to the decision boundary in the feature space the more robust the classifier is to noise. This stands in a line with the conclusion of \cite{fawzi2018analysis}. Another alignment with previous work (e.g.  \cite{bishop1995neural,sokolic2016robust}) is the effect of the norm of $ \w $. Notice that in the proposed theorem, $ \|\w\|_2 $ is multiplied by the noise energy $ \epsilon $ and both have a negative effect on the stability. As a result, one should promote a weight vector of low energy (this is often controlled via a weight decay regularization) as it is capable of increasing the robustness of the sparsity-inspired classification model to noise.

The added value of Sparseland is that a ``well behaved" dictionary, having a small SRIP constant or low mutual coherence, is the key for \emph{stable recovery}, which, in turn, would increase the robustness of the classier. Interestingly, implied from the proof of the obtained results is the fact that \emph{a successful classification can be achieved without recovering the true support} (i.e., the locations of non-zeros in $ \Gama $). This might be counter intuitive, as the support defines the subspace that the signal belongs to. That is, even if the noise in $ \Y $ leads to an estimation $ \hat{\Gama} $ that belongs to slightly different subspace than the one of $ \X $, the input signal could be still classified correctly as long as the dimension of the subspace that it belongs to is small enough (the sparsity constraint).

Our results and perspective on the problem are very different from previous work that studies the robustness to adversarial noise. Fawzi et al. \cite{fawzi2018analysis} suggested a measure for the difficulty of the classification task, where in the linear setting this is defined as the distance between the means of the two classes. Our results differ from these as we heavily rely on a generative model, and so are capable of linking the intrinsic properties of the signal -- its sparsity and filters' design -- to the success of the classification task. This enables us to suggest ways to increase the desired robustness.

A recent work \cite{fawzi2018adversarial} relies on a generative model (similar to ours) that transforms normally-distributed random representation to the signal domain. Their goal was to prove that there exists an upper bound on the robustness that no classifier can exceed. Still, the authors of \cite{fawzi2018adversarial} did not study the effect of the filters nor the network's depth (or the sparsity). Their analysis is very different from ours as we put an emphasis on the stability of sparsity-inspired model and its properties.

As already mentioned, the margin of the data has an impact on the stability as well. Denoting by $ \X' $ a point on the boundary decision, the work reported in \cite{sokolic2016robust} connected the input margin, given by $ \|\X' - \X\|_2 $, to the output distance $ \| f(\Gama') - f(\Gama) \|_2 = \|f(\Gama)\|_2$ through the Jacobian of the classifier $ f(\cdot) $. This connection is of importance as the input margin is directly related to the generalization error. In the scope of this paper, the distance between the signal and its cleaned version in the input space is nothing but the energy of the noise perturbation $ \|\X - \Y\|_2 = \epsilon $. This, in turn, is linked to the score of the classifier distance $ \| f(\hat{\Gama}) - f(\Gama) \|_2 = \| \w^T\hat{\Gama} - \w^T\Gama \|_2 \leq \|\w\|_2\|\hat{\Gama} - \Gama\|_2$ (refer to the proof of Theorem \ref{Thm:StablityRIP} for more details).

We should clarify the term stability used in this section: This refers to \emph{any solver of the pursuit task} that satisfies the following two constraints (i) $ \|\D\Gama-\Y\|_2 \leq \epsilon $, and (ii) $ \|\Gama\|_{0,\infty}^\ss = k $. Later on we shall refer to actual pursuit methods and extend this result further. Concretely, suppose we run the Thresholding pursuit (or the BP) to estimate the representation $ \hat{\Gama} $ of a given $ \Y $. Then, we feed the obtained sparse vector to our linear classifier and hope to assign the input signal to its true label, despite the existence of the adversarial perturbation. Can we guarantee a successful classification in such cases? While this question can be addressed in the CSC case, we shall study the more general multi-layer model. Before doing so, however, we expand the above to the multi-class setting.

\subsection{Multi-class setting}

In order to provide a complete picture on the factors that affect the stability of the classification procedure, we turn to study multi-class linear classifiers. Formally, the discriminant function is described by the following definition: 
\begin{defn}
	A CSC signal $ \X = \D\Gama, \ \|\Gama\|_{0,\infty}^\ss \leq k, $
	is said to belong to class $ y = u $ if the linear discriminant function satisfies $ \forall v \neq u \ \ \ f_u\left(\Gama\right) = \w_u^T\Gama + \omega_u > \w_v^T\Gama + \omega_v = f_v(\Gama), $ where $ u $ stands for the index of the true class, and $ v $ is in the range of $ [1,L] \backslash {v} $. 
\end{defn}
Analogously to the binary case, the decision boundary between class $ u $ and $ v $ is given by $ f_u(\Gama') = f_v(\Gama')  $. Therefore, we formalize the distance to the decision boundary of the class $ y = u $ in the feature space as $ \mathcal{O}_{\mathcal{M}}({\X},y) = \min_{v: v \neq u} f_u\left(\Gama\right) - f_v\left(\Gama\right), $ which measures the distance between the classification result of the $ u $-classifier to the rest $ L-1 $ ones for a given point $ \X $. Similarly to the binary setting, we obtain the minimal distance over all the classes and examples by 
\begin{align}
\mathcal{O}_{\mathcal{M}}^* = \underset{\{\X^j,y^j\}_{j}}{\text{min}} \ \mathcal{O}_{\mathcal{M}}({\X^j},y^j),
\end{align}
where we assume that $ \mathcal{O}_{\mathcal{M}}({\X^j},y^j) > 0 $. Notice that this assumption aligns with the practice, as the common setting in CNN-based classification assumes that a perfect fit of the data during training is possible.
Another measure that will be found useful for our analysis is the distance between the weight vectors. Put formally, we define the multi-class weight matrix $ \W $ of size $ mN \times L $ as $ 
\W = \left[ \w_1 ; \w_2 ; \cdots ; \w_L \right], $
which stores the weight vectors as its columns. The following measure quantifies the mutual Euclidean distance between the classifiers, given by
\begin{align}
\phi(\W) = \underset{u \neq v}{\text{max}} \ {\|\w_u - \w_v\|_2}.
\end{align}
The analogous of this measure in the the binary classification, when $ L = 2 $, is the norm of the classifier being $\|\w\|_2 $, as in this case one can define $ \w_1 = - \w_2 = \frac{1}{2}\w $.

\begin{thm}{(Stable Multi-Class Classification of the CSC Model):} \label{Thm:StablityRIPMC}
	Suppose we are given a CSC signal $ \X = \D \Gama$, $ \|\Gama\|_{0,\infty}^\ss \leq k $, contaminated with perturbation $ \E $ to create the signal $ \Y = \X + \E $, such that $ \|\E\|_{2} \leq \epsilon $. Suppose further that $ f_u(\Gama) = \w_u^T\Gama + \omega_u $ correctly assigns $ \X $ to class $ y = u $.	
	Suppose further that $ \mathcal{O}_{\mathcal{M}}^* > 0 $, and denote by $\hat{\Gama}$ the solution of the $ \Poie $ problem. Assuming that $
	\delta_{2k} < 1 - \left(\frac{2\phi(\W)\epsilon}{\mathcal{O}_{\mathcal{M}}^*}\right)^2, $
	then $ \Y $ will be assigned to the correct class.
	
	Considering the more conservative bound that relies on $ \mu(\D) $ and assuming that
	\begin{equation*}
    	\| \Gama \|_{0,\infty}^{\ss} < k = \frac{1}{2} \left(1 + \frac{1}{\mu(\D)}\left[1 - \left(\frac{2\phi(\W)\epsilon}{\mathcal{O}_{\mathcal{M}}^*}\right)^2\right]\right),
	\end{equation*}
	then $ \Y $ will be classified correctly.
\end{thm}

As one might predict, the same ingredients as in Theorem \ref{Thm:StablityRIP} (coherence of $ \D $ or its SRIP) play a key role here as well. Moreover, in the two-class setting $ \phi(\W) = \|\w\|_2 $, and so the two theorems align. The difference becomes apparent for $ L > 2 $, when the mutual Euclidean distance between the different classifiers influences the robustness of the system to noise. In the context of multi class support vector machine, it was shown \cite{bredensteiner1999multicategory} that the separation margin between the classes $ u $ and $ v $ is $ 2 / \|\w_u - \w_v\|_2$. This observation motivated the authors of \cite{bredensteiner1999multicategory} to minimize the distance $ \|\w_u - \w_v\|_2, \forall u\neq v$ during the training phase. Interestingly, this quantity serves our bound as well.
%This result is intuitive since if the linear classifiers are close to each other (small distance), their ability to distinguish between the different classes is expected to be sensitive to perturbations in the representation vectors.
%In the training phase, one might promote $ \phi(\W) $ to be small as a way to increase the classifier's stability, as suggested in the context of multi class support vector machine \cite{bredensteiner1999multicategory,guermeur2002combining}.
Notice that our theorem also reveals the effect of the number of classes on the robustness. Since $ \phi(\W) $ measures the maximal distance between the $ L $ weight vectors, it is a monotonically increasing function of $ L $ and thereby stiffening our conditions for a successful classification. This phenomenon was observed in practice, indicating that it is easier to ``fool'' the classifier when the number of classes is large, compared to a binary setting \cite{fawzi2018adversarial}.

\section{Robustness bounds for CNN}
\label{sec:MLCSC}

We now turn to extend the above results to a multi-layer setting, and this way shed light on the stability of classic CNN architectures. For simplicity we return to the binary setting, as we have seen that the treatment of multiple classes has a similar analysis. We commence by defining the model we aim to analyze in this part:
\begin{defn} \label{def:binaryDeepClean}
	An ML-CSC signal $\X$ (see definition \ref{Def:DCP})
	is said to belong to class $ y = 1 $ when the linear discriminant function 
	$ f\left(\X\right) = \w^T\Gama_K + \omega $ satisfies $ f(\Gama_K) > 0 $, and $ y = -1 $ otherwise.
\end{defn}
\noindent Notice that the classifier operates on the representation of the last layer, and so the definition of the signal-dependent score $ \mathcal{O}_{\mathcal{B}}(\X,y) $ and the universal $ \mathcal{O}_{\mathcal{B}}^* $ are similar to the ones defined in Section \ref{sec:Two-Class}.
We now turn to the noisy regime, where the adversarial perturbation kicks in:
\begin{defn}\label{Thm:MLClassification}
	For a corrupted ML-CSC signal $\Y = \X + \E$ with a true label $ y $, convolutional dictionaries $ \{\D_i\}_{i=1}^K $, a perturbation energy $ \epsilon $, sparsity levels $ \lamda $, and a classifier $ (\w, \omega) $, define the multi-layer binary classification algorithm as: 
	\begin{align*}
	\text{find} \ \{\hat{\Gama}\}_{i=1}^{K} \ \text{by solving the } \DCPE \ \text{problem};\\ \text{and assign } y = sign\left(f(\hat{\Gama}_K)\right), \qquad \qquad \ \ \
	\end{align*}
	where $ f(\hat{\Gama}_K) = \w^T \hat \Gama_K + \omega $.
\end{defn}

Above, an accurate classification is achieved when $ sign(f(\hat{\Gama}_K)) = sign\left(f({\Gama_K})\right) $. The stability of the multi-layer binary classification algorithm can be analyzed by extending the results of \cite{Papyan2017convolutional} as presented in Section \ref{sec:CSCD}. Therefore, rather than analyzing the properties of the problem, in this section we concentrate on specific algorithms that serve the ML-CSC model -- the L-THR algorithm (i.e. the forward pass of CNN), and its L-BP counterpart. To this end, differently from the previous theorems that we presented, we will assume that the noise is locally bounded (rather than globally\footnote{Locally bounded noise results exist for the CSC as well \cite{Papyan2017working}, and can be leveraged in a similar fashion.}) as suggested in \cite{Papyan2017convolutional}. Put formally, we use the $ \ell_{2,\infty} $-norm to measure the energy of the noise in a vector $ \E $, denoted by $ \|\E\|_{2,\infty}^\pp $, which is defined to be the maximal energy of a $ n_1 $-dimensional patch extracted from it.

\begin{thm}{(Stable Binary Classification of the L-THR):} \label{Thm:StabilityLayeredSoftThresholding}
	Suppose we are given an ML-CSC signal $ \X $ contaminated with perturbation $ \E $ to create the signal $ \Y = \X + \E $, such that $\| \E \|_{2,\infty}^\pp \leq \epsilon_0$. Denote by $|\Gamma_i^{\text{min}}|$ and $|\Gamma_i^{\text{max}}|$ the lowest and highest entries in absolute value in the vector $\Gama_i$, respectively. Suppose further that $ \mathcal{O}_{\mathcal{B}}^* > 0 $ and let $\{\hat{\Gama}_i\}_{i=1}^{K}$ be the set of solutions obtained by running the layered soft thresholding algorithm with thresholds $\{\beta_i\}_{i=1}^{K}$, i.e. $\hat{\Gama}_i=\S_{\beta_i}(\D_i^T\hat{\Gama}_{i-1})$ where $ \S_{\beta_i} $ is the soft thresholding operator and $\hat{\Gama}_{0}=\Y$. Assuming that $\forall \ 1 \leq i \leq K$
	\begin{enumerate} [\quad a) ]
		\item $\| \Gama_i \|_{0,\infty}^\ss < \frac{1}{2} \left( 1 + \frac{1}{\mu(\D_i)} \frac{ |\Gamma_i^{\text{min}}| }{ |\Gamma_i^{\text{max}}| } \right) - \frac{1}{\mu(\D_i)}\frac{ \epsilon_{i-1} }{|\Gamma_i^{\text{max}}|}$;
		\item The threshold $\beta_i$ is chosen according to 
		{\color{\changecolor}
		\begin{multline}
		\label{Eq:ConditionBeta_i}
    		|\Gama_i^{\text{min}}| - C_i - \epsilon_{i-1} 
    		> \beta_i \\
    		> \| \Gama_i \|_{0,\infty}^\ss \mu(\D_i) |\Gama_i^{\text{max}}| + \epsilon_{i-1},
		\end{multline}}
		where
		\begin{equation}
		    \begin{split}
		        C_i&=&( \| \Gama_i \|_{0,\infty}^\ss - 1 ) \mu(\D_i) |\Gama_i^{\text{max}}|, \\
		        \epsilon_i &=& \sqrt{ \| \Gama_{i} \|_{0,\infty}^\pp } \ \Big( \epsilon_{i-1} + C_i + \beta_i \Big);
		    \end{split}
		\end{equation}
		and
		\item $\mathcal{O}_{\mathcal{B}}^* > \|\w\|_2\sqrt{\|\Gama_K\|_0} \Big(\epsilon_{K-1} +C_K + \beta_K\Big)$,	
	\end{enumerate}
	then $ sign(f(\hat{\Gama}_K)) = sign(f(\Gama_K))$.
\end{thm}

\noindent Some of the ingredients of the above theorem are similar to the previous results, but there are several major differences. First, while the discussion in Section \ref{sec:Shallow} concentrated on the stability of the problem, here we get that the forward pass is an algorithm that is capable of recovering the true support of the representations $ \Gama_i $. Still, this perfect recovery does not guarantee a successful classification, as the error in the deepest layer should be smaller than $ \mathcal{O}_{\mathcal{B}}^* $. Second, the forward pass is sensitive to the contrast of the non-zero coefficients in $ \Gama_i $ (refer to the ratio $ { |\Gamma_i^{\text{min}}| } / { |\Gamma_i^{\text{max}}| } $), which is a well known limitation of the thresholding algorithm \cite{Elad_Book}. Third, we see that without a careful regularization (e.g. promoting the coherence to be small) the noise can be easily amplified throughout the layers ($ \epsilon_i $ increases as a function of $ i $). Practitioners refer to this as the error amplification effect \cite{liao2017defense}.

In fact, a similar regularization force is used in Parseval Networks \cite{moustapha2017parseval} to increase the robustness of CNN to adversarial perturbations. These promote the convolutional layers to be (approximately) Parseval tight frames, which are extensions of orthogonal matrices to the non-square case. Specifically, the authors in \cite{moustapha2017parseval} suggested to promote the spectral norm of the weight matrix $ \D_i^T $ to be close to 1, i.e. $ \|\D_i\D_i^T - \I\|_2^2 $, where $ \I $ is the identity matrix. This regularization encourages the average coherence of $ \D_i $ to be small. As our analysis suggests, it was shown that such a regularization significantly improves the robustness of various models to adversarial examples.

Suppose that our data emerges from the ML-CSC model, can we offer an alternative architecture that is inherently better in handling adversarial perturbations? The answer is positive and is given in the form of the L-BP that was suggested and analyzed in \cite{Papyan2017convolutional}. Consider an ML-CSC model of depth two, the LBP algorithm suggests estimating $\Gama_1$ and $\Gama_2$ by solving a cascade of basis pursuit problems. The first stage of this method provides an estimate for $\Gama_1$ by minimizing
\begin{align}
    \hat{\Gama}_1 = \underset{\Gama_1}{\arg\min} \ \|\Y - \D_1\Gama_1\|_2^2 + \xi_1\|\Gama_1\|_1.
\end{align}
Then, an approximation for the deeper representation $\Gama_2$ is given by
\begin{align}
    \hat{\Gama}_2 = \underset{\Gama_2}{\arg\min} \ \|\hat{\Gama}_1 - \D_2\Gama_2\|_2^2 + \xi_2\|\Gama_2\|_1.
\end{align}
Finally, the recovered $\hat{\Gama}_2$ is fed into a classifier, resulting in the predicted label. 

In what follows we build upon the analysis in \cite{Papyan2017convolutional} and show how our theory aligns with the increased stability that was empirically observed by replacing the L-THR algorithm with the L-BP \cite{mahdizadehaghdam2018deep}:
\begin{thm}{(Stable Binary Classification of the L-BP):} \label{Thm:StabilityBP}
	Suppose we are given an ML-CSC signal $\X$ that is contaminated with noise $\E$ to create the signal $\Y = \X + \E$, such that $\| \E \|_{2,\infty}^\pp \leq \epsilon_0$. Suppose further that $ \mathcal{O}_{\mathcal{B}}^* > 0 $, and let $\{\hat{\Gama}_i\}_{i=1}^{K}$ be the set of solutions obtained by running the L-BP algorithm with parameters $\{\xi_i\}_{i=1}^{K}$, formulated as $
	\hat{\Gama}_{i} = \underset{\Gama_i}{\arg\min} \ \xi_i \|\Gama_i\|_1 + \frac{1}{2}\| \D_i\Gama_i - \hat{\Gama}_{i-1} \|_2^2, $
	where $ \hat{\Gama}_{0} = \Y $. Assuming that $\ \forall \  1\leq i\leq K$,
	\begin{enumerate} [\quad a) ]
	\item $\| \Gama_i \|_{0,\infty}^\ss \leq \frac{1}{3} \left( 1 + \frac{1}{\mu(\D_i)} \right)$;
	\item $\xi_i = 4 \epsilon_{i-1}$,\\ where 
	$
	\epsilon_i = \| \E \|_{2,\infty}^\pp \cdot 7.5^i \ \prod_{j=1}^i\sqrt{ \| \Gama_{j} \|_{0,\infty}^\pp };    
	$\\
	and 
	\item $\mathcal{O}_{\mathcal{B}}^* > 7.5\|\w\|_2\sqrt{\|\Gama_K\|_0} \ \epsilon_{K}$,
	\end{enumerate}
	then $ sign(f(\hat{\Gama}_K)) = sign(f(\Gama_K))$.	
\end{thm}
\noindent The proof can be derived by relying on the steps of Theorem \ref{Thm:StabilityLayeredSoftThresholding}, combined with Theorem 12 \textcolor{\changecolor}{ from \cite{Papyan2017convolutional}}. Note that the conditions for the stable classification of the L-BP are not influenced by the ratio $ { |\Gama_i^{\text{min}}| } / { |\Gama_i^{\text{max}}| } $. Moreover, the condition on the cardinality of the representations in the L-BP case is less strict than the one of the L-THR. As such, while the computational complexity of the BP algorithm is higher than the thresholding one, the former is expected to be more stable than the latter. This theoretical statement is supported in practice \cite{mahdizadehaghdam2018deep}. Note that both methods suffer from a similar problem -- the noise is propagated thorough the layers. A possible future direction to alleviate this effect could be to harness the projection (onto the ML-CSC model) algorithm \cite{sulam2017multi}, whose bound is not cumulative across the layers. 

%=======================================================================================================================
%=======================================================================================================================

\section{Numerical experiments}
\label{sec:numerical}
Our study of the stability to bounded noise, in particular Theorems \ref{Thm:StabilityLayeredSoftThresholding} and \ref{Thm:StabilityBP}, introduces a better guarantee for the L-BP, when compared to the well-known L-THR architecture (=CNN). In this section, we aim to numerically corroborate these findings by exploring the actual robustness to adversarial noise of these two architectures. {\color{\changecolor} We achieve this by introducing two sets of experiments: (i) We start with a toy example using synthetically generated data, and show the actual behavior of the L-THR and the L-BP versus their theoretical bounds; and (ii) We proceed by testing these two architectures and exploring their robustness to adversarial noise on actual data, exposing the superiority of the L-BP.}

As described in \cite{Papyan2017convolutional,aberdam2018multi,sulam2018multi} \textcolor{\changecolor}{ and depicted in Figure \ref{Fig:LBP}}, the L-BP is implemented by unfolding the projected gradient steps of the iterative thresholding algorithm. By setting the number of unfolding iterations to zero, the L-BP becomes equivalent to the L-THR architecture.
{\color{\changecolor} Note that both pursuit methods contain the same number of filters, and those are of the same dimensions. Therefore, the same number of free parameters govern both their computational paths. Nonetheless, more unfoldings in the L-BP lead to a higher computational complexity when compared to L-THR.

\subsection{Synthetic experiments}\label{subsec:toy_example}

We start our experimental section with a \emph{toy} example using synthetically generated data, where we have complete access to the generative model and its parameters. This allows us to (i) Compute the theoretical bounds on the permitted noise; and (ii) Compare these predictions with an evaluation of the practical behavior. Our emphasis in this part is on the single layer Thresholding and Basis-Pursuit classifiers, as synthesizing signals from the multi-layered model is far more challenging. Our goal is to show the differences between the theoretical bounds and the measured empirical stability. For completeness, we include experiments with both an undercomplete (having less atoms than the signal dimension) and overcomplete (where the dictionary is redundant) dictionaries.

As already mentioned, we consider the bounds from Theorems \ref{Thm:StabilityLayeredSoftThresholding} and \ref{Thm:StabilityBP} with $K=1$, corresponding to a one hidden layer neural network with a non convolutional (fully connected) dictionary. The two following corollaries describe the bounds of the L-THR and the L-BP in such simplified case.

\begin{corollary}[Stability of one hidden layer L-THR]
    Suppose that $\Y=\D\Gama +\E$, where $\D$ is a general dictionary with normalized atoms, $\norm{\E}_{2}\leq \epsilon$, and $ (\w, \omega) $ is the linear classifier. Suppose also that
    \begin{equation}
        \norm{\Gama}_0\leq \frac{1}{2} \left(1+\frac{\abs{\Gama_{\min}}}{\abs{\Gama_{\max}}}\frac{1}{\mu(\D)}\right)-\frac{\epsilon}{\mu(\D)\abs{\Gama_{\max}}},
    \end{equation}
    and that the threshold $\beta$ set to satisfy:
    \begin{multline} \label{eq:thr_bound_beta_condition}
        \norm{\Gama}_0\mu(\D)\abs{\Gama_{\max}}+\epsilon < \beta < \\ \abs{\Gama_{\min}}-(\norm{\Gama}_0-1)\mu(\D)\abs{\Gama_{\max}}-\epsilon.
    \end{multline}
    Then, the support of $\hat{\Gama}^{\text{THR}}$ is contained in the support of $\Gama$, and
    \begin{equation}
        \norm{\hat{\Gama}^{\text{THR}}-\Gama}_2 \leq \sqrt{\norm{\Gama}_0} \left( \epsilon+ (\norm{\Gama}_0-1)\mu(\D)\abs{\Gama_{\max}} + \beta \right).
    \end{equation}
    Therefore, as long as
    \begin{equation}\label{eq:thr_bound_epsilon_condition}
        \epsilon <\frac{\mathcal{O}_{\mathcal{B}}}{\sqrt{\norm{\Gama}_0}\|\w\|_2}-(\norm{\Gama}_0-1)\mu(\D)\abs{\Gama_{\max}} - \beta ,
    \end{equation}
    the classification is accurate, i.e., $ sign(f(\hat{\Gama}^{\text{THR}})) = sign(f(\Gama))$.
\end{corollary}

\begin{corollary}[Stability of one hidden layer L-BP]
    Suppose that $\Y=\D\Gama+\E$, where $\D$ is a general dictionary with normalized atoms, $\norm{\E}_{2}\leq \epsilon$, and $ (\w, \omega) $ is a linear classifier. Suppose also that $\norm{\Gama}_0\leq \frac{1}{3} \left(1+\frac{1}{\mu(\D)}\right)$,
    and that the Lagrangian multiplier is set to  $\xi=4\epsilon$. Then, the support of $\hat{\Gama}^{\text{BP}}$ is contained in the support of $\Gama$, and
    $\norm{\hat{\Gama}^{\text{BP}}-\Gama}_2 \leq 7.5 \epsilon$.
    Therefore, as long as
    \begin{equation}\label{eq:bp_bound_epsilon_condition}
        \epsilon <\frac{\mathcal{O}_{\mathcal{B}}}{7.5\norm{\Gama}_0\|\w\|_2},
    \end{equation}
    the classification is accurate, i.e., $~sign(f(\hat{\Gama}^{\text{BP}})) = sign(f(\Gama))$.
\end{corollary}

\begin{figure}
    \centering
    \begin{subfigure}{0.5\textwidth}
            \includegraphics[width=0.95\textwidth]{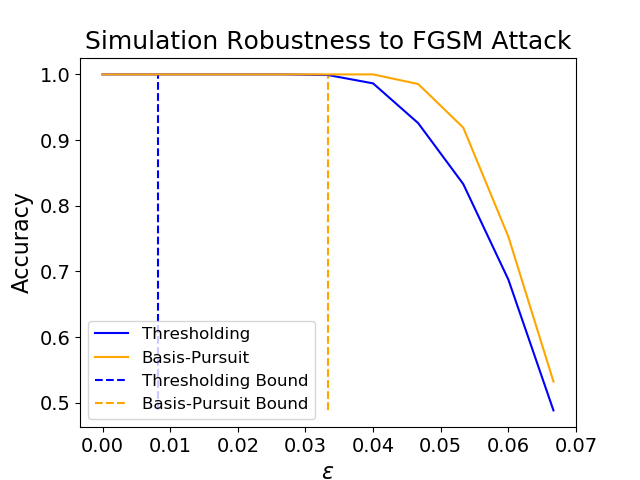}
        \caption{\color{\changecolor}Simulation with an undercomplete dictionary.}
        \label{fig:toy_example_undercomplete}
    \end{subfigure}
    \begin{subfigure}{0.5\textwidth}
        \includegraphics[width=0.95\textwidth]{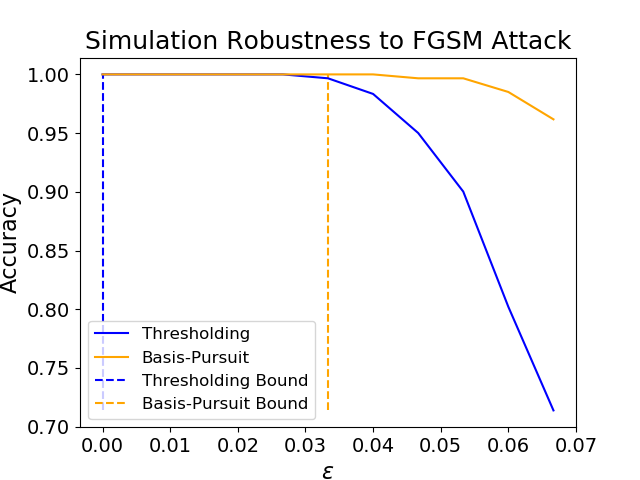}
        \caption{\color{\changecolor}Simulation with an overcomplete dictionary.}
        \label{fig:toy_example_overcomplete}
    \end{subfigure}
    \caption{\color{\changecolor}Accuracy of the THR and the BP versus adversarial noise level, computed on synthetic data. Dashed lines: theoretical bounds; solid lines: empirical performance.}
    \label{fig:toy_examples}
\end{figure}

Figure \ref{fig:toy_examples} presents the theoretical bounds on the adversarial noise amplitude $\epsilon$ for the Thresholding (Equation \eqref{eq:thr_bound_epsilon_condition}) and the Basis-Pursuit (Equation \eqref{eq:bp_bound_epsilon_condition}) classifiers in dash lines. It also shows the empirical stability to adversarial noise under the FGSM (Fast Gradient Sign Method) attack \cite{goodfellow2014explaining}. In these simulations we generate a random normalized and 
unbiased ($\omega=0$) classifier $\w$, and a random dictionary with normalized atoms and with a low mutual-coherence. Then, we randomly produce sparse representations with four nonzeros in the range $\left[|\Gama^{\text{min}}|, |\Gama^{\text{max}}|\right]=\left[1,2\right]$. In order to create a margin $\mathcal{O}_{\mathcal{B}}$ of 1, we project $\Gama$ on the classifier $\w$ and keep only the representations satisfying $\abs{\w^T\Gama} \geq \mathcal{O}_{\mathcal{B}}=1$. Figure \ref{fig:toy_example_undercomplete} presents the undercomplete case with $\D\in \mathbb{R}^{100\times40}$. One can draw three important conclusions from this result: 1) The theoretical stability bound for the BP is better than the THR one; 2) the empirical performance of the BP and THR align with the theoretical predictions; and 3) the bounds are not tight due to the worst-case assumptions used in our work. Note that in this experiment the performance of the two methods is quite close -- this is due to very low mutual-coherence of the chosen dictionary. 

This motivates the next experiment, in which we examine a more challenging setting that relies on an overcomplete dictionary. Figure \ref{fig:toy_example_overcomplete} demonstrates this case with $\D\in \mathbb{R}^{100\times 150}$ and with representations having the same properties as before. In this case, the Thresholding bound collapses to zero as the mutual coherence is too high, and as can be seen, the practical difference between the THR and the BP classifiers becomes apparent.}

\subsection{Real data experiments}

{\color{\changecolor}The goal of the following set of experiments is to show that moving from the traditional feed-forward network (i.e., L-THR) to L-BP can potentially improve stability, not only for simulated data (where the dictionaries and the signals are generated exactly to meet the theorem conditions), but also for real data such as MNIST and CIFAR images. Note that one could wonder whether these images we are about to work with belong to the (possibly multi-layered) sparse representation model. Our approach for answering this question is to impose the model on the data and see how the eventual pursuit (such as the forward-pass) performs. This line of reasoning stands behind many papers that took the sparse representation model (or any of its many variants) and deployed it to true data in order to address various applications.

The networks we are about to experiment with are obtained as unfoldings of the L-THR (Figure \ref{Fig:LTHR}) and the L-BP (Figure \ref{Fig:LBP}) pursuit algorithms, and each is trained in a supervised fashion using back-propagation for best classification performance. Our tested architectures are relatively simple and use a small number of parameters in order to isolate the effect of their differences \cite{sulam2018multi,bibi2018deep}.}

Ideally, in order to demonstrate Theorems \ref{Thm:StabilityLayeredSoftThresholding} and \ref{Thm:StabilityBP}, one should require that the same set of dictionaries is used by the two architectures, in a way that fits our multi-layered model assumptions. However, such setup leads to various difficulties. First, as obtaining these dictionaries calls for training, we should decide on the loss to use. Trained for representation error, these architectures would lead to inferior classification performance that would render our conclusions irrelevant. The alternative of training for classification accuracy would lead to two very different sets of dictionaries, violating the above desire. In addition, as we know from the analysis in \cite{gregor2010learning}, the learned dictionaries are strongly effected by the finite and small number of unfoldings of the pursuit. In the experiments we report hereafter we chose to let each architecture (e.g. pursuit) to learn the best set of dictionaries for its classification result. 

Given the two pre-trained networks, our experiments evaluate the stability by designing noise attacks using the Fast Gradient Sign Method (FGSM) \cite{goodfellow2014explaining} with an increasing amplitude $\epsilon$.
We preform this evaluation on three popular datasets -- MNIST \cite{lecun2010mnist}, CIFAR-10 \cite{krizhevsky2014cifar} and CIFAR-100 \cite{krizhevsky2014cifar}.
For the MNIST case, we construct an ML-CSC model composed of 3 convolutional layers with 64, 128 and 512 filters, respectively, and kernel sizes of $6\times 6$, $6\times 6$ and $4\times 4$, respectively, with stride of 2 in the first two layers. In addition, the output of the ML-CSC model is followed by a fully-connected layer producing the final estimate.
Training is done with the Stochastic Gradient Descent (SGD), with a mini-batch size of 64 samples, learning rate of 0.005 and a momentum weight of 0.9. We decrease the learning rate ten-fold every 30 epochs.

For CIFAR-10 and CIFAR-100 we define an ML-CSC model as having 3 convolutional layers with 32, 64 and 128 filters, respectively, and kernel sizes of $4\times 4$ with stride of 2. In addition, we used a classifier function as a CNN with 4 layers where the first 3 layers are convolutional and the last layer is fully-connected. This effectively results in a 7 layers architecture, out of which the first three are unfolded in the context of the L-BP scheme. As before, all models are trained with SGD and with a decreasing learning rate.

\begin{figure}
    \begin{subfigure}{.45\textwidth}
    \centering
    		\includegraphics[width=\textwidth]{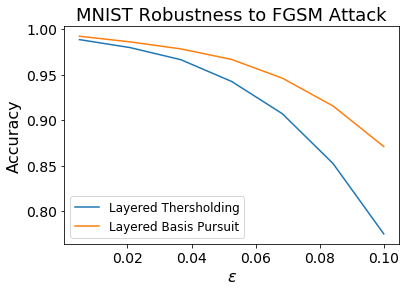}
    		\caption{}
    		\label{fig:comparison_mnist}
    \end{subfigure}
    \begin{subfigure}{.45\textwidth}
    \centering
    		\includegraphics[width=\textwidth]{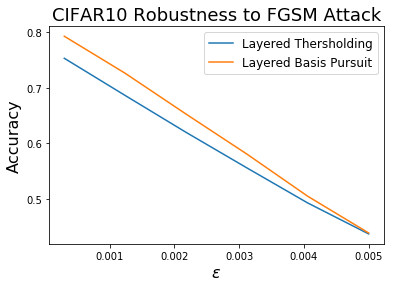}
    		\caption{}
    		\label{fig:comparison_cifar10}
    \end{subfigure}
        \begin{subfigure}{.45\textwidth}
    \centering
    		\includegraphics[width=\textwidth]{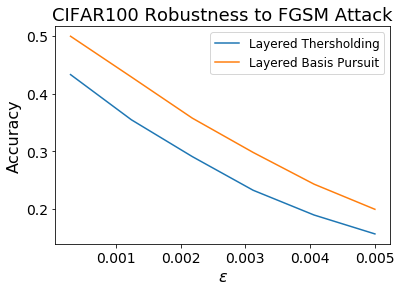}
    		\caption{}
    		\label{fig:comparison_cifar100}
    \end{subfigure}
    \centering
    \caption{Comparison of Layered Thresholding (L-THR) and Layered Basis Pursuit (L-BP) schemes under FGSM attack on (a) MNIST, (b) CIFAR-10 and (c) CIFAR-100 datasets.}
    \label{fig:fgsm}
\end{figure}

Figure \ref{fig:fgsm} presents the results for the two architectures and the three datasets. It is clear that the L-BP scheme is consistently more robust to adversarial interference. This evidence is in agreement with the theoretical results we introduced earlier, suggesting that the L-THR is more sensitive to bounded noise. We note again, however, that the theoretical guarantees presented earlier are not fully applicable here as the dictionaries of each model are different and as some of the assumptions are violated. For example, the minimal distance between classes $\mathcal{O}_{\mathcal{M}}^*$ is not guaranteed to be nontrivial in real images scenario. However, these experiments do support our earlier analysis about the superiority of the L-BP to handle noise attacks.

\section{Conclusions}
\label{sec:conclusions}

This paper presents a general theory for the classification robustness when handling sparsity-modeled signals. In the context of CSC, we studied the stability of the classification problem to adversarial perturbations both for the binary- and multi-class settings. Then, we analyzed the stability of a classifier that operates on signals that belong to the ML-CSC model, which was recently shown to be tightly connected to CNN. This leads to a novel analysis of the sensitivity of the classic forward pass algorithm to adversarial perturbations and ways to mitigate its vulnerability (which was empirically validated in \cite{moustapha2017parseval}). Next, we showed that by relying on the BP algorithm, one can theoretically improve the robustness to such perturbations, a phenomenon that was observed in practice \cite{mahdizadehaghdam2018deep}.

The bounds obtained are all referring to the case where the dictionaries $ \{\D_i\}_{i=1}^K $ and the classification weights $ \{(\w_u,\omega_u)\}_{u=1}^L $ are perfectly known, and thus learning is not covered by this theory. As such, the margin for making an error in our work considers only two prime forces. First, the separability of the data, as manifested by $ \mathcal{O}_{\mathcal{B}}^* $ (or $ \mathcal{O}_{\mathcal{M}}^* $). Second, the chance that our estimated $ \Gama $ deviates from its true value. This can happen due to noise in the input (getting $ \Y $ instead of $ \X $), and/or limitation of the pursuit algorithm. Further work is required in order to bring into account distortions in $ \Gama $ caused by an imperfect estimate of $ \D $'s and $ \w $'s -- this way considering the learning phase as well.

\newtheorem{innercustomgeneric}{\customgenericname}
\providecommand{\customgenericname}{}
\newcommand{\newcustomtheorem}[2]{%
	\newenvironment{#1}[1]
	{%
		\renewcommand\customgenericname{#2}%
		\renewcommand\theinnercustomgeneric{##1}%
		\innercustomgeneric
	}
	{\endinnercustomgeneric}
}

\newcustomtheorem{thmnum}{Theorem}

\begin{appendices}
	
\section{Proof of Theorem 5: stable binary classification of the CSC model}

\begin{thmnum}{5}{(Stable Binary Classification of the CSC Model):}
	Suppose we are given a CSC signal $ \X $, $ \|\Gama\|_{0,\infty}^\ss \leq k $, contaminated with perturbation $ \E $ to create the signal $ \Y = \X + \E $, such that $ \|\E\|_{2} \leq \epsilon $. Suppose further that $ \mathcal{O}_{\mathcal{B}}^* > 0 $ and denote by $\hat{\Gama}$ the solution of the $ \Poie $ problem. Assuming that
	$ \delta_{2k} < 1 - \left(\frac{2{\left\|\w\right\|_2}\epsilon}{\mathcal{O}_{\mathcal{B}}^*}\right)^2, $
	then $ sign(f(\X)) = sign(f(\Y))$.
	
	Considering the more conservative bound that relies on $ \mu(\D) $, and assuming that
	\begin{equation}
		\| \Gama \|_{0,\infty}^{\ss} < k = \frac{1}{2} \left(1 + \frac{1}{\mu(\D)}\left[1 - \left(\frac{2{\left\|\w\right\|_2}\epsilon}{\mathcal{O}_{\mathcal{B}}^*}\right)^2\right]\right),
	\end{equation} then $ sign(f(\X)) = sign(f(\Y))$.  
\end{thmnum}
\begin{proof}
	Without loss of generality, consider the case where $ \w^T\Gama + \omega > 0 $, i.e. the original signal $ \X $ is assigned to class $ y = 1 $. Our goal is to show that $ \w^T\hat{\Gama} + \omega > 0 $. We start by manipulating the latter expression as follows:
	\begin{align}
		\w^T\hat{\Gama} + \omega & = \w^T\left(\Gama + \hat{\Gama} - \Gama \right) + \omega \\
		& = \left(\w^T\Gama + \omega\right) + \w^T\left(\hat{\Gama} - \Gama\right) \\
		& \geq \left(\w^T\Gama + \omega\right) - \left| \w^T\left(\hat{\Gama} - \Gama\right) \right| \\
		& \geq \left(\w^T\Gama + \omega\right) - \left\|\w^T\right\|_2 \left\|\hat{\Gama} - \Gama \right\|_2, \label{eq:classifier}
	\end{align}
	where the first inequality relies on the relation $ a + b \geq a - |b| $ for $ a > 0 $, and the last derivation leans on the Cauchy-Schwarz inequality. Using the SRIP \cite{Papyan2017working} and the fact that both $ \|\Y - \D\Gama\|_2 \leq \epsilon $ and $ \|\Y - \D\hat{\Gama}\|_2 \leq \epsilon $, we get
	\begin{align}
		(1-\delta_{2k}) \|\hat{\Gama} - \Gama \|_2^2 & \leq \|\D\hat{\Gama} - \D\Gama \|_2^2 \leq  4\epsilon^2.
	\end{align}
	Thus, 
	\begin{align}
	    \|\hat{\Gama} - \Gama \|_2^2 & \leq \frac{4\epsilon^2}{1-\delta_{2k}}.
	\end{align}
	Combining the above with Equation \eqref{eq:classifier} leads to (recall that y = 1):
	\begin{align}
		\mathcal{O}_{\mathcal{B}}(\Y,y) = \w^T\hat{\Gama} + \omega & \geq {\w^T\Gama + \omega} - {\left\|\w\right\|_2}\frac{2\epsilon}{\sqrt{1-\delta_{2k}}}.
	\end{align}
	Using the definition of the score of our classifier, satisfying 
	\begin{align}
		0 < \mathcal{O}_{\mathcal{B}}(\X,y) = {\w^T\Gama + \omega}
	\end{align}
	we get
	\begin{align}
		\mathcal{O}_{\mathcal{B}}(\Y,y) \geq \mathcal{O}_{\mathcal{B}}(\X,y) - {\left\|\w\right\|_2}\frac{2\epsilon}{\sqrt{1-\delta_{2k}}}.
	\end{align}
	We are now after the condition for $ \mathcal{O}_{\mathcal{B}}(\Y,y) > 0$, and so we require:
	\begin{align}
		0 & < \mathcal{O}_{\mathcal{B}}(\X,y) - {\left\|\w\right\|_2}\frac{2\epsilon}{\sqrt{1-\delta_{2k}}}\\
		& \leq \mathcal{O}_{\mathcal{B}}^* - {\left\|\w\right\|_2}\frac{2\epsilon}{\sqrt{1-\delta_{2k}}}.
	\end{align}
	where we relied on the fact that $ \mathcal{O}_{\mathcal{B}}(\X,y) \geq \mathcal{O}_{\mathcal{B}}^* $. The above inequality leads to 
	\begin{align}
		\delta_{2k} & < 1 - \left(\frac{2{\left\|\w\right\|_2}\epsilon}{\mathcal{O}_{\mathcal{B}}^*}\right)^2. \label{eq:sucessRIP}
	\end{align}
	
	Next we turn to develop the condition that relies on $ \mu(\D) $. We shall use the relation between the SRIP and the mutual coherence \cite{Papyan2017working}, given by $ \delta_{2k} \geq (2k-1)\mu(\D) $ for all $ k < \frac{1}{2} \left(1 + \frac{1}{\mu(\D)}\right) $.
	Plugging this bound into Equation \eqref{eq:sucessRIP} results in
	\begin{align}
		0 & < \mathcal{O}_{\mathcal{B}}^* - \frac{2 \|\w\|_2 \epsilon}{\sqrt{1-(2k - 1)\mu(\D)}},
	\end{align}
	which completes our proof.	
\end{proof}

\section{Proof of Theorem 7: stable multi-class classification of the CSC model}
\begin{thmnum}{7}{(Stable Multi-Class Classification of the CSC Model):}
	Suppose we are given a CSC signal $ \X $, $ \|\Gama\|_{0,\infty}^\ss \leq k $, contaminated with perturbation $ \E $ to create the signal $ \Y = \X + \E $, such that $ \|\E\|_{2} \leq \epsilon $. Suppose further that $ f_u(\X) = \w_u^T\Gama + \omega_u $ correctly assigns $ \X $ to class $ y = u $.	
	Suppose further that $ \mathcal{O}_{\mathcal{M}}^* > 0 $, and denote by $\hat{\Gama}$ the solution of the $ \Poe $ problem. Assuming that $
	\delta_{2k} < 1 - \left(\frac{2\phi(\W)\epsilon}{\mathcal{O}_{\mathcal{M}}^*}\right)^2, $
	then $ \Y $ will be assigned to the correct class.
	
	Considering the more conservative bound that relies on $ \mu(\D) $ and assuming that
	\begin{equation}
	    \| \Gama \|_{0,\infty}^{\ss} < k = \frac{1}{2} \left(1 + \frac{1}{\mu(\D)}\left[1 - \left(\frac{2\phi(\W)\epsilon}{\mathcal{O}_{\mathcal{M}}^*}\right)^2\right]\right),
	\end{equation}		
	then $ \Y $ will be assigned to the correct class.
\end{thmnum}
\begin{proof}
	Given that $ f_u(\Gama) = \w_u^T\Gama + \omega_u > f_v(\Gama) = \w_v^T\Gama + \omega_v $ for all $ v \neq u $, i.e. $ \X $ belongs to class $ y = u $, we shall prove that $ f_u(\hat{\Gama}) > f_v(\hat{\Gama}) $ for all $ v \neq u $. Denoting $ \Delta = \hat{\Gama} - \Gama $, we bound from below the difference $ f_u(\hat{\Gama}) - f_v(\hat{\Gama})$  as follows:
	\begin{multline} \label{eq:classifierMulti}
		\left[\w_u^T\hat{\Gama} + \omega_u\right] - \left[\w_v^T\hat{\Gama} + \omega_v\right] \\
		\begin{split}
		& = \left[\w_u^T\left(\Gama + \Delta \right) + \omega_u \right]- \left[\w_v^T\left(\Gama + \Delta \right) + \omega_v\right]\\
		& = \left[\w_u^T\Gama + \omega_u\right] - \left[\w_v^T\Gama + \omega_v\right] + \left(\w_u^T - \w_v^T\right) \Delta \\
		& \geq f_u(\Gama) - f_v(\Gama) - \left| \left(\w_u^T - \w_v^T\right) \Delta \right| \\
		& \geq f_u(\Gama) - f_v(\Gama) - \|\w_u^T - \w_v^T\|_2\|\Delta\|_2
		.\end{split}
	\end{multline}
	Similarly to the proof of Theorem \ref{Thm:StablityRIP}, the first inequality holds since $ a + b \geq a - |b| $ for $ a = f_u(\Gama) - f_v(\Gama) > 0 $, and the last inequality relies on the Cauchy-Schwarz formula. Relying on $ \phi(\W) $ that satisfies
	\begin{align}
		\phi(\W) \geq \|\w_u - \w_v\|_2,
	\end{align}
	and plugging $\|\Delta \|_2^2 \leq \frac{4\epsilon^2}{1-\delta_{2k}} $ into Equation \eqref{eq:classifierMulti}  we get
	\begin{align}
		f_u(\hat{\Gama}) - f_v(\hat{\Gama}) & \geq f_u(\Gama) - f_v(\Gama) - \phi(\W) \frac{2\epsilon}{\sqrt{1-\delta_{2k}}} \\
		& \geq \mathcal{O}_{\mathcal{M}}({\X},y) - \phi(\W) \frac{2\epsilon}{\sqrt{1-\delta_{2k}}} \\
		& \geq \mathcal{O}_{\mathcal{M}}^* - \phi(\W) \frac{2\epsilon}{\sqrt{1-\delta_{2k}}},	
	\end{align}
	where the second to last inequality holds since  $ f_u(\Gama) - f_v(\Gama) \geq  \mathcal{O}_{\mathcal{M}}({\X},y)$, and the last inequality follows the definition of $ \mathcal{O}_{\mathcal{M}}^* $. As such, we shall seek for the following inequality to hold:
	\begin{align}
		0 & < \mathcal{O}_{\mathcal{M}}^* - \phi(\W)\frac{2\epsilon}{\sqrt{1-\delta_{2k}}} \\ 
		\rightarrow \delta_{2k} & < 1 - \left(\frac{2\phi(\W)\epsilon}{\mathcal{O}_{\mathcal{M}}^*}\right)^2.
	\end{align}
	Similarly to the binary setting, one can readily write the above in terms of $ \mu(\D) $.
\end{proof}

\section{Proof of Theorem 10: stable binary classification of the L-THR}
\begin{thmnum}{10}{(Stable Binary Classification of the L-THR):}
	Suppose we are given an ML-CSC signal $ \X $ contaminated with perturbation $ \E $ to create the signal $ \Y = \X + \E $, such that $\| \E \|_{2,\infty}^\pp \leq \epsilon_0$. Denote by $|\Gamma_i^{\text{min}}|$ and $|\Gamma_i^{\text{max}}|$ the lowest and highest entries in absolute value in the vector $\Gama_i$, respectively. Suppose further that $ \mathcal{O}_{\mathcal{B}}^* > 0 $ and let $\{\hat{\Gama}_i\}_{i=1}^{K}$ be the set of solutions obtained by running the layered soft thresholding algorithm with thresholds $\{\beta_i\}_{i=1}^{K}$, i.e. $\hat{\Gama}_i=\S_{\beta_i}(\D_i^T\hat{\Gama}_{i-1})$ where $ \S_{\beta_i} $ is the soft thresholding operator and $\hat{\Gama}_{0}=\Y$. Assuming that $\forall \ 1 \leq i \leq K$
	\begin{enumerate} [\quad a) ]
		\item $\| \Gama_i \|_{0,\infty}^\ss < \frac{1}{2} \left( 1 + \frac{1}{\mu(\D_i)} \frac{ |\Gamma_i^{\text{min}}| }{ |\Gamma_i^{\text{max}}| } \right) - \frac{1}{\mu(\D_i)}\frac{ \epsilon_{i-1} }{|\Gamma_i^{\text{max}}|}$;
		\item The threshold $\beta_i$ is chosen according to 
		\begin{equation}
    		|\Gama_i^{\text{min}}| - C_i - \epsilon_{i-1} 
    		> \beta_i
    		> K_i + \epsilon_{i-1},
		\end{equation}
		where
		\begin{equation}
		    \begin{split}
		        C_i&=&( \| \Gama_i \|_{0,\infty}^\ss - 1 ) \mu(\D_i) |\Gama_i^{\text{max}}|, \\
		        K_i& =& \| \Gama_i \|_{0,\infty}^\ss \mu(\D_i) |\Gama_i^{\text{max}}|, \\
		        \epsilon_i &=& \sqrt{ \| \Gama_{i} \|_{0,\infty}^\pp } \ \Big( \epsilon_{i-1} + C_i + \beta_i \Big);
		    \end{split}
		\end{equation}
		and
		\item $\mathcal{O}_{\mathcal{B}}^* > \|\w\|_2\sqrt{\|\Gama_K\|_0} \Big(\epsilon_{K-1} +C_K + \beta_K\Big)$,	
	\end{enumerate}
	then $ sign(f(\Y)) = sign(f(\X))$.
\end{thmnum}
\begin{proof}
	Following Theorem 10 in \cite{Papyan2017working}, if
	assumptions (a)--(c) above hold $\forall \ 1 \leq i \leq K$ then
	\begin{enumerate}
		\item The support of the solution $\hat{\Gama}_i$ is equal to that of $\Gama_i$; and \label{Item:SuccForwardPass}
		\item $\| \Gama_i - \hat{\Gama}_i \|_{2,\infty}^\pp \leq \epsilon_i$, where $\epsilon_i$ defined above.
	\end{enumerate}
	In particular, the last layer satisfies 
	\begin{equation}
		\| \Gama_K - \hat{\Gama}_K \|_{\infty} \leq \epsilon_{K-1} + C_K + \beta_K. \label{Eq:ErrorLastLayer}
	\end{equation}
	Defining $ \Delta = \hat{\Gama}_K - \Gama_K $, we get
	\begin{align}
		\|\Delta\|_{2} \leq \|\Delta\|_\infty\sqrt{\|\Delta\|_0} = \|\Delta\|_\infty\sqrt{\|\Gama_K\|_0},
	\end{align}
	where the last equality relies on the successful recovery of the support. Having the upper bound on $ \|\Delta\|_2 $, one can follow the transition from Equation \eqref{eq:classifier} to Equation \eqref{eq:sucessRIP} (see the proof of Theorem \ref{Thm:StablityRIP}), leading to the following requirement for accurate classification:
	\begin{align}
		\mathcal{O}_{\mathcal{B}}^* - \|\w\|_2\|\Delta\|_\infty\sqrt{\|\Gama_K\|_0} > 0.
	\end{align}
	Plugging Equation \eqref{Eq:ErrorLastLayer} to the above expression results in the additional condition that ties the propagated error throughout the layers to the output margin, given by
	\begin{align}
		\mathcal{O}_{\mathcal{B}}^* > \|\w\|_2\sqrt{\|\Gama_K\|_0} \Big(\epsilon_{K-1} + C_K + \beta_K\Big).
	\end{align}
\end{proof}

\end{appendices}

\bibliographystyle{plain}
\bibliography{MyBib}

\begin{thebibliography}{10}

\bibitem{aberdam2018multi}
Aviad Aberdam, Jeremias Sulam, and Michael Elad.
\newblock Multi-layer sparse coding: The holistic way.
\newblock {\em SIAM Journal on Mathematics of Data Science}, 1(1):46--77, 2019.

\bibitem{bibi2018deep}
Adel Bibi, Bernard Ghanem, Vladlen Koltun, and Rene Ranftl.
\newblock Deep layers as stochastic solvers.
\newblock In {\em International Conference on Learning Representations}, 2019.

\bibitem{bishop1995neural}
Chris Bishop.
\newblock {\em Neural networks for pattern recognition}.
\newblock Oxford university press, 1995.

\bibitem{bredensteiner1999multicategory}
Erin~J Bredensteiner and Kristin~P Bennett.
\newblock Multicategory classification by support vector machines.
\newblock In {\em Computational Optimization}, pages 53--79. Springer, 1999.

\bibitem{candes2008restricted}
Emmanuel~J Candes.
\newblock The restricted isometry property and its implications for compressed
  sensing.
\newblock {\em Comptes rendus mathematique}, 346(9-10):589--592, 2008.

\bibitem{Elad_Book}
Michael Elad.
\newblock {\em Sparse and Redundant Representations: From Theory to
  Applications in Signal and Image Processing}.
\newblock Springer Publishing Company, Incorporated, 1st edition, 2010.

\bibitem{fawzi2018adversarial}
Alhussein Fawzi, Hamza Fawzi, and Omar Fawzi.
\newblock Adversarial vulnerability for any classifier.
\newblock {\em arXiv preprint arXiv:1802.08686}, 2018.

\bibitem{fawzi2018analysis}
Alhussein Fawzi, Omar Fawzi, and Pascal Frossard.
\newblock Analysis of classifiers’ robustness to adversarial perturbations.
\newblock {\em Machine Learning}, 107(3):481--508, 2018.

\bibitem{goodfellow2016deep}
Ian Goodfellow, Yoshua Bengio, Aaron Courville, and Yoshua Bengio.
\newblock {\em Deep learning}, volume~1.
\newblock MIT press Cambridge, 2016.

\bibitem{goodfellow2014explaining}
Ian~J Goodfellow, Jonathon Shlens, and Christian Szegedy.
\newblock Explaining and harnessing adversarial examples.
\newblock {\em ICLR}, 2015.

\bibitem{gregor2010learning}
Karol Gregor and Yann LeCun.
\newblock Learning fast approximations of sparse coding.
\newblock In {\em Proceedings of the 27th International Conference on Machine
  Learning (ICML-10)}, pages 399--406, 2010.

\bibitem{krizhevsky2014cifar}
Alex Krizhevsky, Vinod Nair, and Geoffrey Hinton.
\newblock The cifar-10 dataset.
\newblock {\em online: http://www. cs. toronto. edu/kriz/cifar. html}, 2014.

\bibitem{kurakin2016adversarial}
Alexey Kurakin, Ian Goodfellow, and Samy Bengio.
\newblock Adversarial examples in the physical world.
\newblock {\em arXiv preprint arXiv:1607.02533}, 2016.

\bibitem{lecun2015deep}
Yann LeCun, Yoshua Bengio, and Geoffrey Hinton.
\newblock Deep learning.
\newblock {\em Nature}, 521(7553):436--444, 2015.

\bibitem{lecun2010mnist}
Yann LeCun, Corinna Cortes, and CJ~Burges.
\newblock Mnist handwritten digit database.
\newblock {\em AT\&T Labs [Online]. Available: http://yann. lecun.
  com/exdb/mnist}, 2, 2010.

\bibitem{liao2017defense}
Fangzhou Liao, Ming Liang, Yinpeng Dong, Tianyu Pang, Jun Zhu, and Xiaolin Hu.
\newblock Defense against adversarial attacks using high-level representation
  guided denoiser.
\newblock {\em IEEE-CVPR}, 2018.

\bibitem{liu2016delving}
Yanpei Liu, Xinyun Chen, Chang Liu, and Dawn Song.
\newblock Delving into transferable adversarial examples and black-box attacks.
\newblock {\em ICLR}, 2017.

\bibitem{mahdizadehaghdam2018deep}
Shahin Mahdizadehaghdam, Ashkan Panahi, Hamid Krim, and Liyi Dai.
\newblock Deep dictionary learning: A parametric network approach.
\newblock {\em arXiv preprint arXiv:1803.04022}, 2018.

\bibitem{Mairal2014}
Julien Mairal, Francis Bach, and Jean Ponce.
\newblock Sparse modeling for image and vision processing.
\newblock {\em arXiv preprint arXiv:1411.3230}, 2014.

\bibitem{moustapha2017parseval}
Cisse Moustapha, Bojanowski Piotr, Grave Edouard, Dauphin Yann, and Usunier
  Nicolas.
\newblock Parseval networks: Improving robustness to adversarial examples.
\newblock {\em ICML}, 2017.

\bibitem{Papyan2017convolutional}
Vardan Papyan, Yaniv Romano, and Michael Elad.
\newblock Convolutional neural networks analyzed via convolutional sparse
  coding.
\newblock {\em Journal of Machine Learning Research}, 18(83):1--52, 2017.

\bibitem{Papyan2017working}
Vardan Papyan, Jeremias Sulam, and Michael Elad.
\newblock Working locally thinking globally: Theoretical guarantees for
  convolutional sparse coding.
\newblock {\em IEEE Trans. on Sig. Proc.}, 65(21):5687--5701, 2017.

\bibitem{sokolic2016robust}
Jure Sokoli{\'c}, Raja Giryes, Guillermo Sapiro, and Miguel~RD Rodrigues.
\newblock Robust large margin deep neural networks.
\newblock {\em IEEE Trans. on Sig. Proc.}, 65(16):4265--4280, 2016.

\bibitem{sulam2018multi}
Jeremias Sulam, Aviad Aberdam, Amir Beck, and Michael Elad.
\newblock On multi-layer basis pursuit, efficient algorithms and convolutional
  neural networks.
\newblock {\em IEEE transactions on pattern analysis and machine intelligence},
  2019.

\bibitem{sulam2017multi}
Jeremias Sulam, Vardan Papyan, Yaniv Romano, and Michael Elad.
\newblock Multilayer convolutional sparse modeling: Pursuit and dictionary
  learning.
\newblock {\em IEEE Transactions on Signal Processing}, 66(15):4090--4104,
  2018.

\bibitem{szegedy2013intriguing}
Christian Szegedy, Wojciech Zaremba, Ilya Sutskever, Joan Bruna, Dumitru Erhan,
  Ian Goodfellow, and Rob Fergus.
\newblock Intriguing properties of neural networks.
\newblock {\em ICLR}, 2014.

\bibitem{Zeiler2010}
Matthew~D Zeiler, Dilip Krishnan, Graham~W Taylor, and Rob Fergus.
\newblock Deconvolutional networks.
\newblock In {\em IEEE-CVPR}, 2010.

\end{thebibliography}

\end{document}